\documentclass[letterpaper, 10 pt, conference]{ieeeconf}  

\IEEEoverridecommandlockouts                              

\usepackage{array,multirow}
\usepackage{graphicx}
\usepackage[bookmarks=false]{hyperref}
\usepackage[noadjust]{cite}

\title{\LARGE \bf
Prioritized Multi-agent Path Finding for Differential Drive Robots}

\author{Konstantin Yakovlev$^{1}$, Anton Andreychuk$^{2}$ and Vitaly Vorobyev$^{3}$
\thanks{$^{1}$Konstantin Yakovlev is with Federal Research Center for Computer Science and Control of RAS and with Higher School of Economics.}%
\thanks{$^{2}$Anton Andreychuk is with Peoples' Friendship University of Russia (RUDN University).}%
\thanks{$^{3}$Vitaly Vorobyev is with Kurchatov Institute.
\newline Corresponding author is Konstantin Yakovlev, {\tt\small{yakovlev@isa.ru}}.
\newline This is a preprint of the paper accepted to ECMR 2019: https://ieeexplore.ieee.org/document/8870957%
}
}

\begin{document}

\maketitle
\thispagestyle{empty}
\pagestyle{empty}

\begin{abstract}

Methods for centralized planning of the collision-free trajectories for a fleet of mobile robots typically solve the discretized version of the problem and rely on numerous simplifying assumptions, e.g. moves of uniform duration, cardinal only translations, equal speed and size of the robots etc., thus the resultant plans can not always be directly executed by the real robotic systems. To mitigate this issue we suggest a set of modifications to the prominent prioritized planner -- AA-SIPP(m) -- aimed at lifting the most restrictive assumptions (syncronized translation only moves, equal size and speed of the robots) and at providing robustness to the solutions. We evaluate the suggested algorithm in simulation and on differential drive robots in typical lab environment (indoor polygon with external video-based navigation system). The results of the evaluation provide a clear evidence that the algorithm scales well to large number of robots (up to hundreds in simulation) and is able to produce solutions that are safely executed by the robots prone to imperfect trajectory following. The video of the experiments can be found at \url{https://youtu.be/Fer_irn4BG0}.

\end{abstract}

\section{INTRODUCTION}

Problem of finding feasible, collision free trajectories for multiple robots navigating in a shared environment is a challenging problem that is lacking general efficient solution. 
Essentially, two approaches to multi-robot navigation are common. One is to adjust the velocity profiles of the robots in a reactive fashion, taking into account current observations. Methods implementing this approach, e.g. ORCA \cite{van2011reciprocal}, typically scale well to large number of robots but can not guarantee that each robot reaches its goal. Another approach is to plan the collision-free trajectories beforehand assuming that robots will execute them precisely or within some given tolerance that has been accounted for (see works \cite{le2017cooperative, cap2015a, Wagner2011, bartak2018multi} etc.). In this work we adopt the planning approach.

Multi-robot planners typically solve a discretized version of the problem -- multi-robot path planning on graphs \cite{yu2016optimal} also attributed as multi-agent path finding (MAPF) \cite{ma2017ai}. They often provide guarantees on completeness, optimality (or bounded sub-optimality) of the solutions (w.r.t to discretization). Unfortunately most of the MAPF planners, like the ones presented in \cite{standley2010, de2013push, sharon2015, barer2014}, rely heavily on numerous simplifying assumptions, e.g. neglecting agents' size, assuming all agents move synchronously with the same speed etc. To make the MAPF solutions applicable to real robots one can post-process them as proposed in \cite{honig2016multi} or can modify the planning algorithm itself in the way that it lifts as much constraints as possible \cite{walker2018icts, honig2018trajectory, bartak2018scheduling, li2019multi, ma2019lifelong} and/or produces robust solutions that are more likely to be  executed safely \cite{vcap2016provably, wagner2017path, atzmon2018robust}.

\begin{figure}
    \centering
    \includegraphics[width=0.5\textwidth]{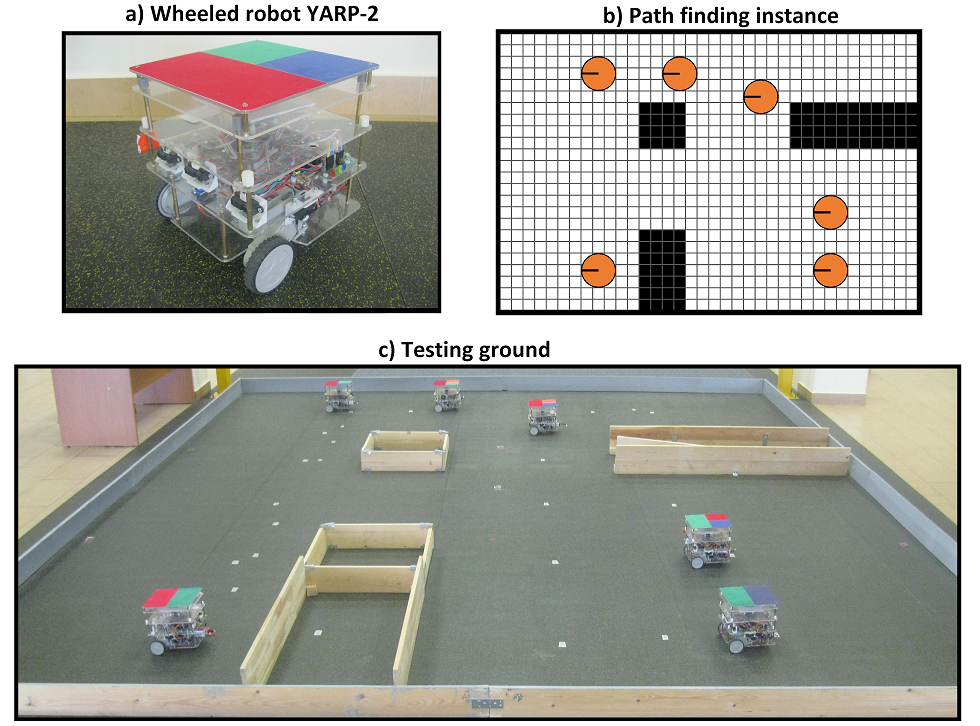}
    \caption{Motivation scenario. A fleet of the wheeled robots (a) having to safely navigate the shared environment (c). The discretized model of the problem is shown in (b).}
    \label{fig_environemnt}
\end{figure}

Following this line of research we suggest the prioritized multi-agent planner, particularly suited for differential drive robots that have a shape of (or can be modelled as) disks. It builds up on our previous work on any-angle safe interval path planning \cite{yakovlev2017aasipp} and, to the best of our knowledge, it is the first multi-agent planner that \textit{in practice} \textit{i}) does not restrict robots' moves to syncronized translations; \textit{ii}) allows moves of arbitrary durations (i.e. durations that are not strictly tied to the preliminary discretized timeline); \textit{iii}) allows planning for robots with different sizes; \textit{iv}) does not require moving speed to be the same; \textit{v}) takes rotation actions into account when planning. It also preemptively tries to minimize the risk of collisions, occurring due to imperfect execution. On top of that, to increase the chance of finding the solution and to decrease the cost of the solution, as it is known that prioritized planning is not complete/optimal in general,
we incorporate such techniques as deterministic re-scheduling and start safe intervals into the algorithm.

The proposed algorithm is extensively tested in simulation and on real robots. Results of the evaluation provide a strong evidence that algorithm scales well to large number of robots (up to hundreds in simulation) and that the solutions it produces can be safely executed by the robots with imperfect localization and trajectory following.

\section{PROBLEM STATEMENT}
Consider $n$ robots populating the workspace that is tessellated to a grid (see Fig. \ref{fig_environemnt}b). Each robot can translate, rotate or wait-in-place. Waiting and rotating is allowed only at the center of a grid cell and initially all robots are waiting at their start cells. When moving inertial effects are neglected and the speed is fixed for a move (but can vary from one move to the other). The safety zone of a robot (or a robot itself) is modeled as an open-disk of radius $r^{(i)}$. 

A state (configuration) for a robot is a tuple $(x, y, \theta)$, where $x, y$ are the spatial coordinates and $\theta$ is the heading angle. The configuration is valid w.r.t to static obstacles if the distance between $(x, y)$ and the closest point of the obstacle is greater or equal $r^{(i)}$. The trajectory for a robot $i$ is a mapping $tr^{(i)}: [0, \infty) \rightarrow C^{(i)}_{free}$, s.t. $tr^{(i)}(0)=\mathbf{start}$ and $\exists T_i \; \forall t \geq T_i \; tr^{(i)}(t)=\mathbf{goal}$. Here $\mathbf{start} = (x_s, y_s, \theta_s)$ and $\mathbf{goal} = (x_g, y_g, \theta_g)$ are the initial and the goal states and $C^{(i)}_{free}$ is the union of all feasible states for the robot $i$. The cost of the trajectory is the time needed execute it.

Two trajectories are said to be collision free if the robots following them never collide. The problem is to find $n$ trajectories for the robots, s.t. each pair of them is collision free. The cost of the solution might be either the makespan, that is the maximum over the costs of individual trajectories, or the flowtime, that is the sum of costs. In this work optimal solutions are not targeted, but the low cost solutions are, obviously, preferable.

\section{METHOD}
\subsection{Prioritized planning}
Prioritized planning is a well-known approach \cite{erdmann1987}, when each robot is assigned a unique priority and then individual trajectories are planned sequentially in accordance with the imposed ordering.
Prioritized planning is complete in case individual planner avoids start locations of the lower-priority robots and the instance satisfies certain conditions \cite{cap2015a}. In some cases prioritized planning is also optimal \cite{ma2018searching}. In general, though, it is neither optimal nor complete. One of the approaches to mitigate this issue is to re-plan with other priority ordering in case of failure. In \cite{bennewitz2001optimizing} random re-shuffling was proposed, in \cite{ma2018searching} a deterministic algorithm for systematic exploration of priority orderings was suggested. In this work we adopt a heuristic algorithm proposed in \cite{andreychuk2018two} for re-assigning priorities. In case of failure it sets the priority of the failed robot to maximum and re-plans. This approach is fast and easy to implement and
in practice it significantly raises the chances of finding solution and outperforms random re-ordering. 

Another enhancement for prioritized planning we implemented is utilizing start safe intervals \cite{andreychuk2018two} (SSIs). Planning with SSIs means that start locations of the low-priority robots are considered to be blocked for the predefined amount of time and un-blocked afterwards. Such an ad-hoc technique contributes to increasing the chance of finding a solution, thus decreasing the number of re-planning attempts, thus lowering down the runtime.

For individual planning we use the enhanced AA-SIPP algorithm. Original planner \cite{yakovlev2017aasipp} assumed that all agents are of equal radii and move with the same speed. The planning was conducted for translations (and waits) only. We lift these assumptions, i.e. we plan for translations and rotations (and waits), assume that different robots might have different rotation/translations speeds and take them into account when planning. We also perform the computation of the so-called earliest arrival time in a more straightforward fashion compared to a verbose algorithm described in \cite{yakovlev2017aasipp}. As a result we end up with a more versatile and easy-to-implement version of AA-SIPP.

The source code of the resultant planner is open and available at \url{https://github.com/PathPlanning/AA-SIPP-m}.

\subsection{Individual planner: enhanced AA-SIPP}

\subsubsection{High-level overview}
AA-SIPP stands for any-angle safe interval path planning. It is a heuristic search planner that groups contiguous, collision-free time points for each element of the configuration space into the intervals and use them to define the nodes of the search space \cite{phillips2011}. Utilizing intervals relieves the search effort as now for each configuration only a limited number of search nodes, proportional to how many dynamic obstacles hit this configuration, might be generated while a conventional discrete planner might generate $[1, ..., T]$ nodes for a single configuration, where $T$ is the time horizon that might be very large.

To illustrate the idea of interval planning suppose that the robot $cur$ needs to move to the cell $(3, 2)$ as shown on Fig.\ref{fig_intervals}, and it is known that two high-priority robots pass nearby and hit the cell. In this case 3 safe intervals for a configuration $(3, 2, \theta)$ should be considered, i.e. the robot can reach (and stay in) the cell before the first obstacle hits is, or in between the first obstacle leaves away and the second obstacle hits the configuration, or after the second obstacle moves away. SIPP-based planners, e.g AA-SIPP, consider all three possibilities by trying to generate 3 search nodes corresponding to this configuration. Within each safe interval a paradigm of ``reach the configuration as early as possible'' is adopted. Such an approach coupled with with the A* search strategy guarantees finding optimal solution (w.r.t to the discretization), i.e. the trajectory that avoids all the static and dynamic obstacles and minimizes time. Detailed code of the algorithm can be found in \cite{yakovlev2017aasipp}.

\subsubsection{Computing intervals}
AA-SIPP search nodes are identified by the tuples $s=[cfg, interval]$. $cfg=(x, y, \theta)$ accounts for robot's position and heading. $interval=[t_s, t_f]$ -- is the contiguous period of time for a configuration, during which there is no collision and it is in collision one time point prior and one time point after the period.

\begin{figure}
    \centering
    \includegraphics[width=0.45\textwidth]{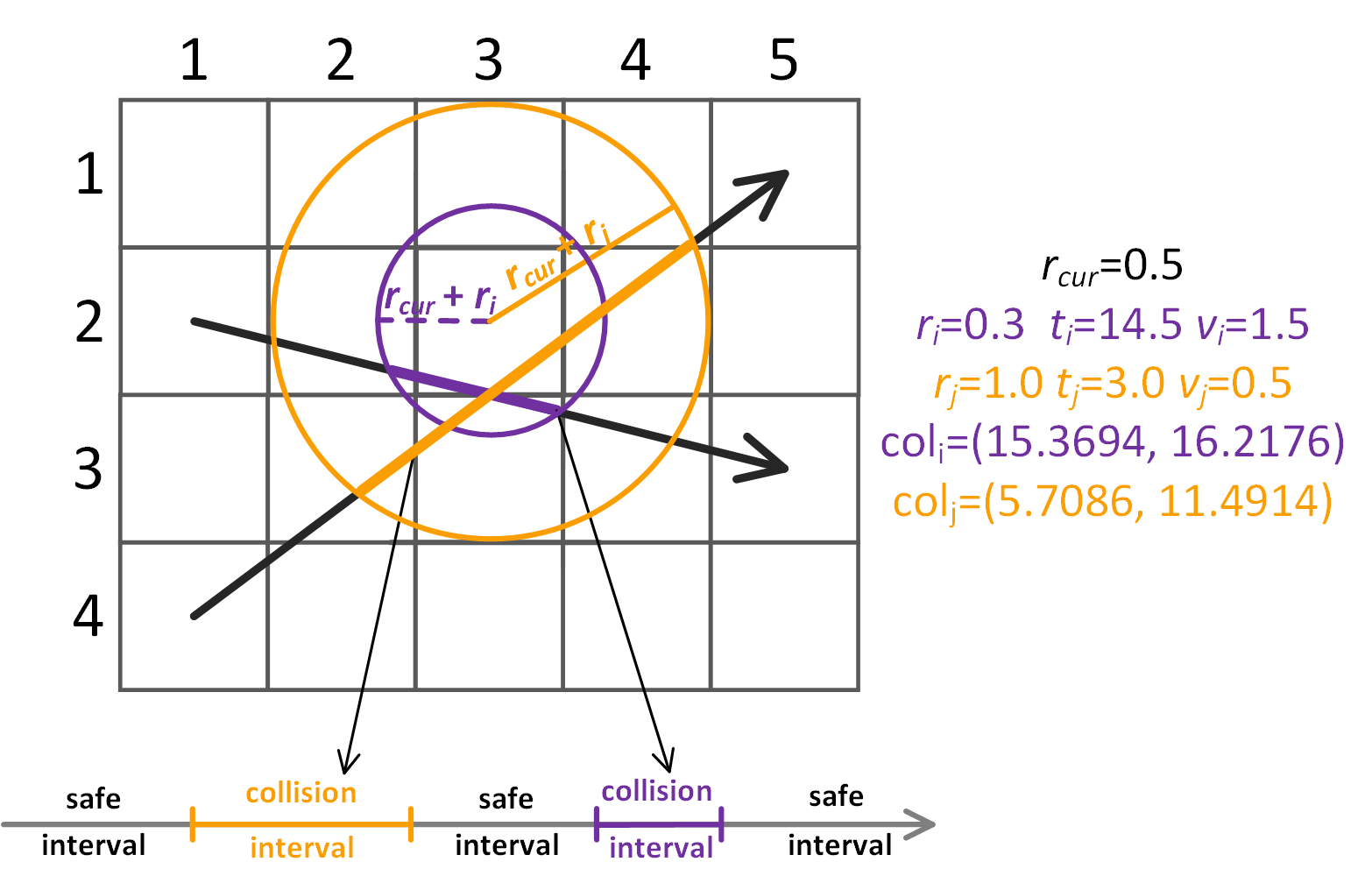}
    \caption{Computing safe and collision intervals for a cell through which passes two different robots.}
    \label{fig_intervals}
\end{figure}

When a planner considers a move to a cell it should attempt generating $k$ successors corresponding to $k$ distinct safe intervals. Figure \ref{fig_intervals} illustrates how safe intervals are computed. First, we draw circumferences centered at that cell of the radii equal to $r^{(cur)} + r^{(i)}$, where $r^{(cur)}$ is the safety radius of the current robot (the one we are planning for) and $r^{(i)}$ is the radii of the moving obstacles (high-priority robots) that pass nearby. Then, using conventional formulas of geometry, we compute the coordinates of the points at which the circumferences intersect the corresponding path segments. Knowing these points, as well at the obstacle trajectories, we may now compute the collision intervals and inverse them to get the safe intervals we are interested in. Now the planner can attempt to generate the successors (one per each safe interval). Whether an attempt succeeds depends on whether the move to the target cell is valid w.r.t static and dynamic obstacles. 

\subsubsection{Estimating the feasibility of the move w.r.t. static obstacles}

Naive approach to estimate the feasibility of the move between the grid cells is to a) assume that robot fits inside the cell, i.e. $r^{(cur)} \leq 0.5l$, where $l$ is the size of the cell and b) constrain the robot to move to 4 cardinal directions only. In that case one can simply check whether the target cell is traversable. We wish not to restrict agent's size, e.g. to be able to handle agents that are bigger then the grid cells, and to handle moves between arbitrary grid cells. To do so we developed the original procedure of estimating the feasibility of the move w.r.t. static obstacles.

\begin{figure}
    \centering
    \includegraphics[width=0.45\textwidth]{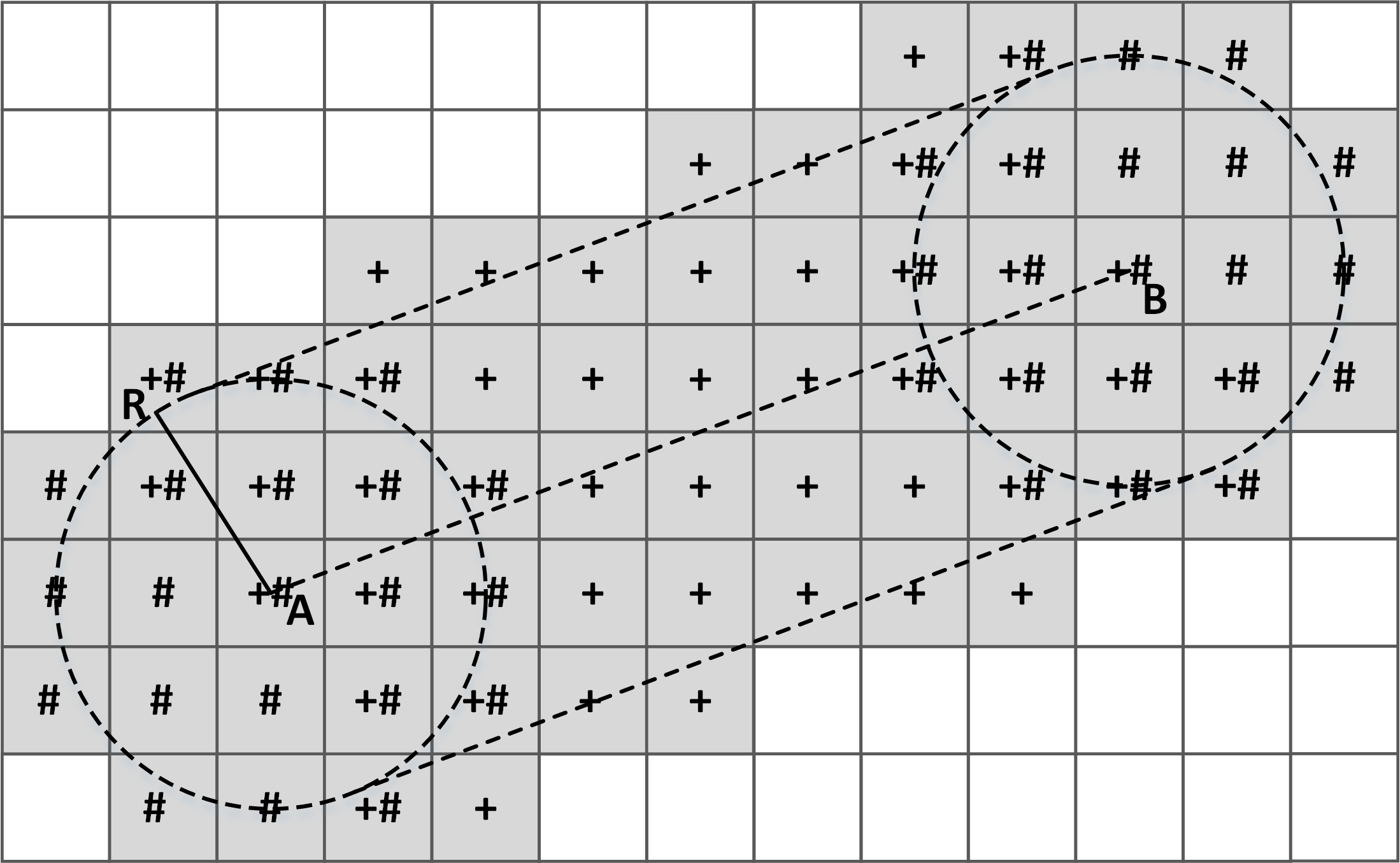}
    \caption{Estimating the feasibility of the translation move for a robot of arbitrary radius.}
    \label{fig_los}
\end{figure}

The idea behind the procedure is to identify which cells are hit by the robot moving along the line connecting the move's endpoints and check their traversability. This is done in a similar way to how algorithms from computer graphics, e.g. Bresenham algorithm \cite{bresenham1965algorithm}, identify pixels that lie along the straight line between two fixed points -- see Fig. \ref{fig_los}. We iteratively process the columns of the grid and for each column compute how many cells residing up/down the line to check (these cells are marked with ``+''). We additionally process the endpoints of the move to identify which cells line inside the circumference of the given radius (these cells are marked with ``\#'') and check them as well.

Instead of estimating the feasibility of the move w.r.t to static obstacles in the described fashion, one might think of a more conventional approach: enlarge obstacles by the half-radius of the agent and treat it as a moving point. The problem with that approach is two-fold. First, we need to make such transformation for each agent of different size, thus we will end up with storing and operating with multiple workspaces. Second, as the disk-robots can be of arbitrary radius, e.g. $r=0.6$ cells, we may end up with marking some cells as blocked only because they partially overlap the no-way zone. Thus, we may fail to find the path although it exists (due to some enlarged obstacles have merged). 

\subsubsection{Estimating the feasibility of the move w.r.t. dynamic obstacles and computing earliest arrival time}

If the translation to the destination cell is feasible, it must be performed as soon as possible, following SIPP's paradigm of reaching each node at the earliest possible time. Problem is that the immediate translation might lead to a collision with a high-priority robot passing nearby or waiting/rotating at the cell that lies on the way. We treat all cases uniformly by considering wait/rotate moves as translations with zero velocity. The only exception is when a high-priority robot is waiting at its goal position. In this case the move for the current robot can not be performed anyhow and the corresponding successor is discarded.

To detect collision between two translating disks (one -- corresponding to the current robot, another -- to the high-priority one) we rely on the closed-loop formula from \cite{guy2015}, which gives yes/no answer to the collision query (it also computes time to collision but we do not use it for planning purposes). This formula takes disks radii and translation velocities as arguments, so we are not restricted to predefined move speeds and agent sizes anymore.

If collision occurs we increment the duration of the wait action preceding the translation on some predefined value $\delta$ and repeat. In such a way we find the time when current robot can safely start moving. This time moment is guaranteed to exist as dynamic obstacle will sooner or later move out of the way. The earliest arrival time is now computed based on the time spent for waiting and for translating. Finally, the successor is generated in case two conditions hold: 1) the departure time belongs to the safe interval of the source node; 2) the arrival time belongs to the safe interval of the target node.

The proposed approach to compute the earliest arrival time is straightforward and can be easily implemented compared to the approach originally introduced in \cite{yakovlev2017aasipp}.

\subsubsection{Handling rotations}
Commonly multi-agent path finding solvers assume that an agent do not need to rotate before translation. In the studied domain this assumption does not hold. Fortunately, as we do not discretize timeline into the timesteps we can naturally plan for rotations of any duration and on any angle needed, by simply assuming that before translating robot spends $(\theta' - \theta)\omega$ time units for rotating, where $\omega$ is rotation speed, $\theta$ is current heading and $\theta'$ is the desired heading. As said before, waits and rotations are treated uniformly when checking for collisions, so, from the collision-avoidance perspective, rotation is equivalent to wait action. Thus, rotation actions of arbitrary duration can be seamlessly embedded to the suggested planning framework reaffirming its versatility.

\subsection{Increasing robustness}\label{increasing_robustness}
When it comes to real robots one can not expect perfect execution, which might lead to collisions although the plan is valid. To increase the robustness of the generated solutions we suggest two approaches. First, one can inflate the robots' size thus introducing extra-safety zone around them. We can do so by virtue of the proposed collision checking routines that are not tailored to specific size. Second, one can add additional wait of some arbitrary duration, say $d$, before any translation move when planning. At the execution phase, in case a robot fails to arrive to the waypoint on time, it has to wait before next move not $d$ but $d - delay$ timepoints, where $delay$ is the amount of time the robot is late. Thus, chances are each translation move actually starts on time and the path following error is discharged (at least partially). 
We evaluated both suggested approaches on real robots and they showed convincing results. More sophisticated approaches, e.g. the one described in \cite{vcap2016provably}, might also be realized within the suggested framework.

\begin{figure*}[t!]
\includegraphics[width=\textwidth]{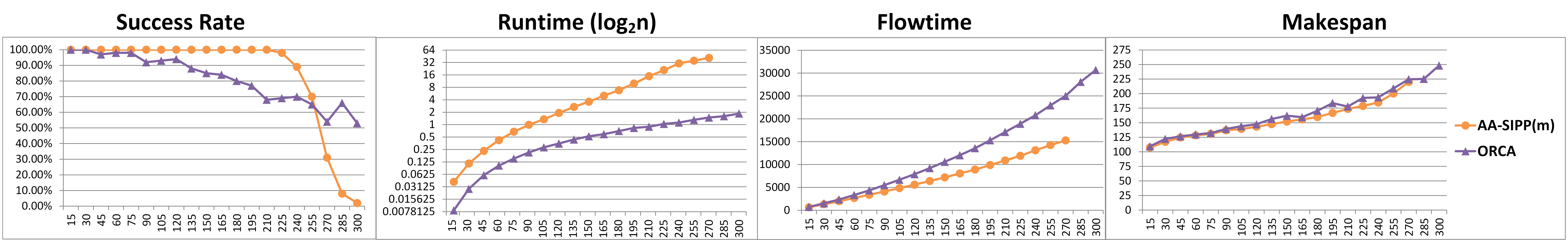}
\caption{Results on the 64x64 empty grid. OX-axis is the number of agents (on all charts). Success rate is in percent, runtime -- in seconds, flowtime/makespan -- in time units.}
\label{fig_res_1}
\end{figure*}

\begin{figure*}[t!]
\includegraphics[width=\textwidth]{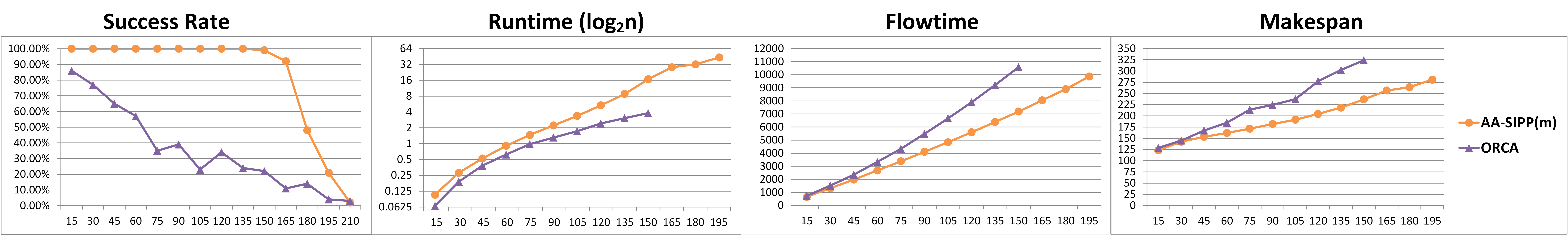}
\caption{Results on the 64x64 grid with 10 rectangular obstacles. OX-axis is the number of agents (on all charts). Success rate is in percent, runtime -- in seconds, flowtime/makespan -- in time units.}
\label{fig_res_2}
\end{figure*}

\section{EXPERIMENTAL EVALUATION}
\subsection{Simulated experiments}
By conducting experiments in simulation we pursued two aims. First, we wanted to assess how the suggested algorithm scales to large number of robots. Second, we wanted to compare it to direct competitors. Unfortunately the second aim is hard achieve as, to the best of authors knowledge, at the moment of writing this paper there existed no other centralized multi-agent path finding solver that \textit{simultaneously} (and without pre- or post-processing) handles rotations and translations into arbitrary direction, supports actions not tied to discrete timeline, supports varying robots' size and speed. At the same time the prominent decentralized algorithm -- ORCA \cite{van2011reciprocal} -- supports (almost) all of those features (ORCA does not support rotate actions explicitly but rather adjusts robots' velocity at a high rate that can be considered analogous to rotation). Thus, we chose it for the comparison. The source code of both algorithms is publicly available\footnote{\url{https://github.com/PathPlanning/AA-SIPP-m}}\textsuperscript{,} \footnote{\url{http://gamma.cs.unc.edu/RVO2/}}. Video of the selected experiments can be found at \url{https://youtu.be/Fer_irn4BG0}.

\subsubsection{Empty hall} 100 meta-instances accommodating 300 robots were generated on empty $64 \times 64$ grid. Three types of robots were involved. Robots of the first type are small and fast: their radius equals 0.3 cells and translation speed is 1.5 cells per time unit. Second type robots translate with the speed of 1 cell per time unit and their radius is 0.5. Robots of the third type are large and slow: their radius equals 1.0 and translation speed is 0.5. Rotation speed is $180^\circ$ per time unit for all robots. Each meta-instance contained 100 robots of each type. Start and goal locations and headings were chosen randomly. During experimental evaluation we transformed each meta-instance to the instance with the specific number of robots. We started with 5 robots of each type, i.e. 15 robots on a map, and then gradually increased the number up to 300 of robots. Time limit was set to 1 minute. If the algorithm was not able to produce a solution within allotted time, we stopped it and count this run as failure.

The following parameters were used for AA-SIPP(m). Robots were assigned initial priorities based on the Euclidean distance between start and goal -- the lower the distance, the higher the priority was. The value for the Start safe interval was set to 3. We used deterministic re-scheduling in case of failure by raising the priority of the failed robot. For ORCA we set time boundary equal to 10, sight radius to 15 and maximum neighbors to 15. These values were chosen empirically based on the preliminary evaluation of AA-SIPP(m) and ORCA.

Resultant metrics are shown on. Fig. \ref{fig_res_1}. We average across the instances that were successfully solved by both algorithms. No results for AA-SIPP(m) for 285 and 300 agents are given due to the low success rate.

Overall, AA-SIPP(m) managed to solve all the instances with up to 210 robots, while ORCA eventually failed even when 45 robots were involved. Densely populated environments posed a problem to AA-SIPP(m). When the number of robots exceeded 240 AA-SIPP(m) almost always required re-scheduling and often the time was out. As a result AA-SIPP(m) solved less than 10\% of instances with 285 or 300 agents while the success rate of ORCA dropped to about 50-60\%.

AA-SIPP(m) always outperforms ORCA in terms of flowtime and almost always in terms of makespan. When the number of robots is low ORCA's flowtime is 10-15\% worse. As the number of agents grows up and reaches 150, the difference becomes more than 50\%. In terms of makespan the difference is not so significant and is about 4\% on average.

\subsubsection{Non-empty hall}To evaluate the algorithms' performance in non-empty environments we added 10 static obstacles to the map. Each obstacle was a rectangle formed of $20 \times 2$ cells. To let ORCA avoid them we used the code, proposed by the algorithm's authors, that builds the visibility graph and finds a reference path for each robot on this graph using Dijkstra's algorithm. 

The results are presented in Fig. \ref{fig_res_2}. In contrast to the previous tests, success rate of ORCA is not 100\% even for 15 agents. In terms of solution cost AA-SIPP(m) shows much better results than ORCA. When the number of robots is low their results are rather close, but when the number of agents exceeds 60 the difference becomes significant and reaches about 2x in flowtime and almost 40\% in makespan.

\subsubsection{Summary}
Principally, AA-SIPP(m) scales well to large number of robots in simulation. When static obstacles are present AA-SIPP(m) significantly outperforms ORCA no matter how many robots are involved. Success rate is higher and the flowtime/makespan is notably lower. The only advantage of ORCA is lower runtime, which is predictable as it is a reactive navigation algorithm based on a rather simple collision-avoidance strategy compared to deliberative planning via heuristic search performed by AA-SIPP(m). When there are no obstacles and the number of robots is moderate AA-SIPP(m) solves more instances than ORCA and provides solutions of better quality. The only case when ORCA can be considered preferable is when the map is empty and the number of robots is very high (more than 285 in our case).

\subsection{Evaluation on the wheeled robots}
We conducted experiments with 6 identical differential drive robots depicted on Fig. \ref{fig_environemnt}a. Each robot is 21x21 cm in size and is able to move with maximum speed of 10 cm per second (that was the speed we used for planning). Rotation speed is $24^\circ$ per second. Each robot is equipped with the colored marker, that is tracked by the external vision-based navigation system composed of the 6 web-cameras, and with a APC220 radio module to communicate with the central computing station. The latter is the PC laptop that runs Ubuntu and ROS. We implemented ROS modules for \textit{i}) retrieving, filtering and processing the video-stream from the cameras; \textit{ii}) localizing the robots, i.e. computing $(x, y, \theta)$ state for each robot; \textit{iii}) communicating with the robots, i.e. sending them the next action of a plan to execute. Action execution is performed locally, i.e. using the controllers installed on the robots. The PID regulator is used to maintain near-constant translation and rotation velocity.

The polygon is depicted on Fig. \ref{fig_environemnt}c. It is 6x4.8 m bounded rectangle containing 3 obstacles. This polygon was represented as $ 108 \times 72$ grid (cell size was 5 cm). For planning purposes robots were modelled as disks. The minimum radius we used was 15 cm which corresponds to almost-zero safety zone for a robot, as when it rotates it describes a circle of radius 14.8 cm.

Before evaluating how well AA-SIPP(m) plans are executed by the robots, we have run a preliminary series of experiments with only one robot involved, aimed at estimating the accuracy of trajectory execution. We executed 2 different trajectories 20 times each and tracked the position of the robot to compare the executed trajectory against the planned one. We discarded the heading and compared only $(x, y)$ components of the trajectories. The average RMSE turned out to be 30.65 cm, that is 1.45 of robot actual size.

The error is quite big so we conducted AA-SIPP(m) evaluation on 6 robots, setting the radius of the disk modelling each robot to be 15, 25 and 35 cm. First, we planned without adding waits, then added 5s waits before each translation (and use these waits to compensate delays of reaching the waypoints). 10 different solvable instances were generated. Thus, in total $10 \times 3 \times 2 = 60$ experimental runs on real robots were made.

\begin{figure}[t]
    \centering
    \includegraphics[width=0.48\textwidth]{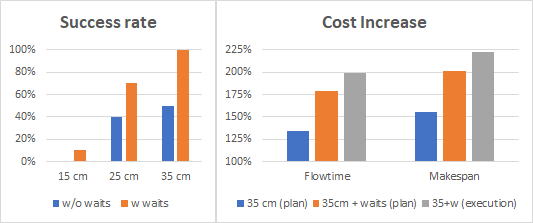}
    \caption{Left: Percent of safely executed instances when planning for different robots' size and with/without wait augmentation technique. Right: flowtime/makespan overhead compared to the ideal plan, i.e. the one that was obtained for radius = 15 cm and without wait augmentation.}
    \label{fig_success_rate}
\end{figure}

The results are depicted on Fig. \ref{fig_success_rate}. Executing trajectories that were planned without inflating the safety zone and without utilizing add-waits technique leads to collisions in 100\% of cases. Inflating the safety zone and using wait augmentation made the produced solutions much more robust and suitable for execution by the robots. In fact, 100\% of plans were safely executed when the radius was 35 cm and 5 s waits were added before each translation. The price one has to pay for such a robustness is a 2x increase in flowtime/makespan when compared to ideal plans, i.e. the ones that were constructed w/o waits for 15 cm disks (see Fig. \ref{fig_success_rate} on the right).

In general, the suggested planner proved to be a flexible and versatile tool in practice. After appropriate tuning it was capable of providing robust solutions that were safely executed by real robots in the absence of perfect localization and path following.

\section{CONCLUSIONS AND FUTURE WORK}
In this work we suggested an enhanced multi-agent path finding algorithm based on prioritization and safe interval path planning that lifts numerous assumptions characteristic to the algorithms of this kind. The resultant planner supports varying robots' size, translation/rotation speeds, non-fixed moves' durations etc. and is particularly suitable for differential drive wheeled robots. One of the directions of future research is increasing the computational efficiency of the algorithm, another one is applying the proposed techniques to the planners that do not rely on prioritization and guarantee completeness/optimality, e.g. CBS-planners. 

\section*{ACKNOWLEDGMENT}
This work was supported by the Russian Foundation for Basic Research (project \#18-37-20032), RUDN University Program 5-100, special program of the presidium of Russian Academy of Sciences

\addtolength{\textheight}{-8cm}   








\bibliographystyle{IEEEtran}
\bibliography{yakovlev.bib}


\end{document}